# CMLM-CSE: Based on Conditional MLM Contrastive Learning for Sentence Embeddings

ZHANG Wei[1,2] and CHEN Xu[1]

[1]Hangzhou Yizhi Intelligent Technology Co., Ltd., HangZhou, ZheJiang, China

chenxu@yiwise.com

[2]School of Computer Science, Zhejiang University, HangZhou, ZheJiang, China

22121040@zju.edu.cn

*Abstract*

*Traditional comparative learning sentence embedding directly uses the encoder to extract sentence features, and then passes in the comparative loss function for learning. However, this method pays too much attention to the sentence body and ignores the influence of some words in the sentence on the sentence semantics. To this end, we propose CMLM-CSE, an unsupervised contrastive learning framework based on conditional MLM. On the basis of traditional contrastive learning, an additional auxiliary network is added to integrate sentence embedding to perform MLM tasks, forcing sentence embedding to learn more masked word information. Finally, when $Bert_{base}$ was used as the pretraining language model, we exceeded SimCSE by 0.55 percentage points on average in textual similarity tasks, and when $Roberta_{base}$ was used as the pretraining language model, we exceeded SimCSE by 0.3 percentage points on average in textual similarity tasks.*

## 1. INTRODUCTION

Learning universal sentence embeddings is a fundamental problem in natural language processing and has been studied extensively in the literature [1, 2, 3, 4, 7]. Much recent work has shown that pre-trained language models fine-tuned by contrast learning on unlabeled datasets can learn a good representation of sentences [2, 3, 4, 7]. Contrastive learning takes a self-supervised approach to training by using multiple data enhancements of its own samples as positive sample pairs and other samples within the same training batch as negative samples, pulling the positive sample pairs closer to the semantic representation space and pushing away the negative sample pairs during the training process.

Chen et al. found that different data enhancements (e.g., random cropping, rotation, random inversion, color dithering, adding Gaussian noise, etc.) play a crucial role in pre-training visual models for contrast learning [5], but these data enhancements are usually unsuccessful when applied to sentence embedding contrast learning.Gao et al. proposed in SimCSE [3] that by a simple dropout-based data expansion approach to construct positive samples is much more effective than complex data expansion approaches based on synonym replacement, word deletion, etc. In hindsight, this is not surprising, since deletion or replacement of sentences usually changes the meaning of the sentence and also shows that pre-trained language models are very sensitive to data augmentation by word replacement. In response, Chuang et al [7] added an equal-variable contrast learning [9] agent task to SimCSE [3] for error judgment of masked words, which was further improved, but because there are only two results for right and wrong, there is a large randomness in the judgment.

In this paper, we combine Bert pre-trained language model and propose a new method of contrast learning sentence embedding based on word features, which improves the semantic information contained in the sentence encoding and further improves the effect of text-semantic similarity matching of sentences by adding an agent task of mask word prediction.

## 2 Related Work

### 2.1 Contrastive Learning

The purpose of contrastive learning is to learn an effective semantic representation by bringing semantically similar pairs of samples closer together and pushing apart pairs of samples that are not semantically similar. Given a sentence pair $\{x_i, x_i^+\}$ where $x_i$ and $x_i^+$ are a pair of semantically identical or similar sentences, we treat these two sentences as positive sample pairs. We adopt the contrastive learning framework proposed by Chen et al. in SimCLR [5], and use a cross-entropy loss function for a training batch [10]. Let $h_i$ and $h_i^+$ represent the features of $x_i$ and $x_i^+$, respectively. A minimal training batch contains N pairs of $(x_i, x_i^+)$ and the specific loss function is as follows:

$$\ell_i = -log \frac{e^{sim(h_i, h_i^+)/\tau}}{\sum_{j=1}^{N} e^{sim(h_i, h_j^+)/\tau}} \quad (1)$$

Where, N represents the number of sentences in a mini-batch, and $\tau$ is a temperature hyperparameter. $\text{sim}(h_1, h_2) = \frac{h_1^T h_2}{||h_1|| \cdot ||h_2||}$ represents the cosine similarity.

## 3 CMLM-CSE model structures

The CMLM-CSE model architecture proposed in this paper, as shown in Figure 1, consists of two parts: the standard SimCSE [3] (left side of Figure 1) and an auxiliary network (right side of Figure 1. Specifically, (1) The text $x$ is input twice into the sentence encoder, and two different sentence encodings $h^+$ and $h$ are obtained through dropout for computing the contrastive loss. (2) The text $x$ is randomly masked to obtain the masked text $x'$, which is then passed through the lexical feature extractor of the auxiliary network to extract the lexical features of the sentence to obtain $h'$. Then, the sentence embedding $h$ and the hidden states of the lexical features of the sentence except for the [CLS] position generated by SimCSE are concatenated to obtain $h''$. Finally, the semantic feature Extractor module predicts the masked token, and the conditional MLM loss is computed.

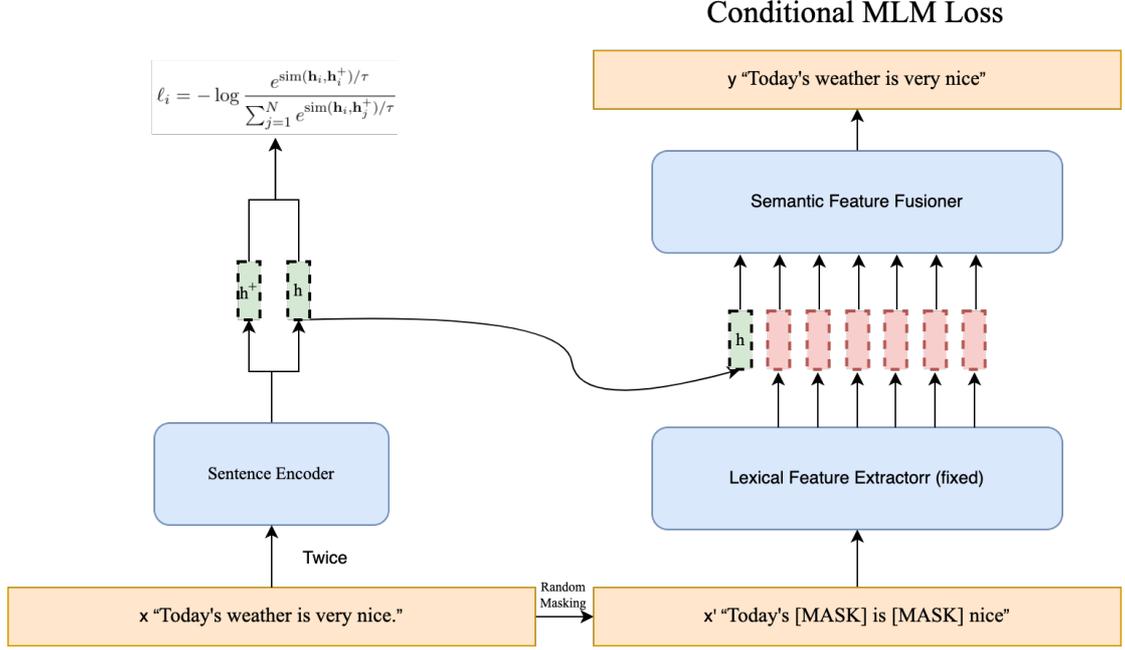

Figure 1: The structure of the CMLM-CSE model.

### 3.1 SimCSE and Contrastive Loss

we adopt the same structure as SimCSE and define a set of sentence samples $X = \{x_i\}_{i=1}^{N}$. In this paper, we use a pre-trained BERT model as the sentence encoder. It is worth noting that the BERT model structure consists of multiple layers of Transformer Block, and each Transformer Block has two dropout layers, which are located after the attention probabilities and fully connected layers, respectively. This ensures that even if two completely identical samples are input at different times, their outputs will still be different. We define $h_i^m = f_\theta(m(x_i))$ where $m(x_i)$ is the operation of randomly dropping out $x_i$. We input the same sample into the encoder twice, and obtain two different feature encodings $h_i^{m_i}$ and $h_i^{m_i'}$ by using two different dropout masks $m$ and $m'$ respectively. Then, we compute the contrastive loss function as follows:

$$l_i = -\log \frac{e^{\text{sim}\left(h_i^{m_i}, h_i^{m_i'}\right)/\tau}}{\sum_{j=1}^{N} e^{\text{sim}\left(h_i^{m_i}, h_j^{m_i'}\right)/\tau}} \quad (4)$$

where $N$ is the length of a training batch, $\tau$ is the temperature, which is a hyperparameter, and $L_{contrast} = \sum_{i=1}^{N} l_i$.

### 3.2 Auxiliary Network and Conditional MLM Task

The auxiliary network consists of two parts: the lexical feature extractor and the semantic feature fusioner. The lexical feature extractor selects the first eight layers of BERT. Ganesh and Jawahar [24] have demonstrated that the intermediate layers of BERT encode rich linguistic information and exhibit the characteristics of lexical features in the bottom layers, syntactic features in the middle layers, and semantic features in the top layers. By leveraging the rich syntactic features in the intermediate layers of the pre-trained BERT model and the sentence vector $h$ with semantic information output by SimCSE, we hope to provide more sufficient information to the semantic feature fusioner.

Furthermore, it is worth emphasizing that the lexical feature extractor is frozen during the parameter training process to avoid the semantic information being learned by the lexical feature extractor. Since the lexical feature extractor is frozen, in order to reduce the conditional MLM loss, it is necessary for the sentence embedding h to contain more semantic information to assist the semantic feature fusioner in correctly predicting the masked tokens. The semantic feature fusioner consists of three Transformer Blocks.

During training, in this part, the input sentence $x_i$ is first randomly masked to obtain $x_i'$. Then, $x_i'$ is input into the lexical feature extractor to obtain the corresponding lexical features $h_i'$. The sentence vector $h_i$ output by SimCSE and the hidden states $h'$ of the lexical features except for the [CLS] position output by the lexical feature extractor are concatenated as follows:

$$h_i'' = [h_i, h_i' > 0]$$

where $h_i'\{> 0\}$ represents the hidden states of other positions except for position 0, and $h_i''$ is the concatenated vector. $h_i''$ is then input into the semantic feature fusioner to obtain the probability distribution of the masked token $z_i$, which is used to calculate the cross-entropy loss as follows:

$$L_{MLM} = -\sum_{i=1}^{N}\sum_{j=1}^{p} x_j^{(i)} \log\left(z_j^{(i)}\right) \qquad (5)$$

where $N$ is the number of samples in a training batch, $p$ is the fixed sentence length selected during training, and $x_j^{(i)}$ represents the j-th $token$ of the i-th sentence in batch $x$ that is not masked.

Compared to the traditional MLM task, the MLM task in this work is conditional, which means restoring the masked token under the premise of having the semantic information of the sentence. If the sentence embedding h carries sufficient semantic information, even with only three Transformer Blocks in the lexical feature extractor, it can still effectively restore the masked token.

### 3.3 Loss Function Combination

The loss function of CMLM-CSE consists of two parts: the contrastive loss of SimCSE and the conditional MLM loss of the auxiliary network.

$$L = L_{contrast} + \lambda L_{MLM} \qquad (6)$$

where $\lambda$ is a hyperparameter that determines the influence of the conditional MLM loss function on the contrastive loss function. Since the contrastive loss is simpler than the conditional MLM loss, in order to balance the two loss terms, the value of $\lambda$ is set to be relatively small. Please refer to Section 4.1.2 for details of the ablation experiment analysis.

# 4 Experimental Settings and Result Analysis

## 4.1 Experimental Settings

In our experiments, we adopted the same settings as the unsupervised SimCSE, and used two pre-trained models, Bert[11] and Roberta[12], to initialize our sentence encoder. The lexical feature extractor is the first eight layers of the Bert pre-trained model, and the semantic feature fusioner consists of three Transformer Blocks. Note that the lexical feature extractor is frozen during the entire training process and does not undergo parameter updates, serving only as an auxiliary learning tool.

## 4.2 Data

For unsupervised pre-training, we selected the Wikipedia dataset provided in the source code of SimCSE[3], which contains one million simple English sentences. For model evaluation, we chose seven semantic textual similarity (STS) datasets for semantic similarity evaluation, including STS 2012-2016[13], STS Benchmark[14], and SICK-Relatedness[6]. All STS experiments are completely unsupervised, which means that no STS training dataset is used, and all sentence embeddings are generated using a fixed model trained in an unsupervised manner.

## 4.3 Result Analysis

### 4.3.1 Comparison of Text Similarity Performance based on Pre-trained Word Embeddings

We compared our model results with SimCSE[3], IS-BERT[16], DeCLUTR[20], CMLM[17], CT-BERT[19], SG-OPT[18], some post-processing methods such as Bert-flow[21] and BERT-whitening[22], and some simple baseline models such as GloVe[23]. The comparison results are shown in Table 1. When selecting $Bert_{base}$ as the pre-trained model, our model outperforms SimCSE by 2.12, 0.51, 0.64, 0.50, and 1.33 percentage points on the STS12, STS14, STS15, STS16, and STS-B datasets, respectively, with an average improvement of 0.55 percentage points over SimCSE. Compared with our own reproduced DiffCSE model, our model outperforms it by 2.09, 1.16, 1.02, and 0.39 percentage points on the STS12, STS14, STS15, and STS16 datasets, respectively, with an average improvement of 0.22 percentage points over DiffCSE. When selecting $RoBERTa_{base}$ as the pre-trained model, our model outperforms SimCSE by 0.49, 0.17, 1.11, 0.49, and 0.2 percentage points on the STS12, STS13, STS14, STS15, and STS-B datasets, respectively, with an average improvement of 0.3 percentage points over SimCSE.

Table 1: Performance (Spearman correlation coefficient) of different sentence embedding models on STS tasks. ∇ results are from [15]; ♡ results are from [3]; ♣ results are from [16]; ♠results are from [17]; ♯ results are from [18]; ∗ results are from our experiments.

| Model | STS12 | STS13 | STS14 | STS15 | STS16 | STS-B | SICK-R | Avg. |
|---|---|---|---|---|---|---|---|---|
| GloVe embeddings(avg.)∇ | 55.14 | 70.66 | 59.73 | 68.25 | 63.66 | 58.02 | 53.76 | 61.32 |

| | | | | | | | |
|---|---|---|---|---|---|---|---|
| BERT$_{base}$ (first-last avg.)♡ | 39.70 | 59.38 | 49.67 | 66.03 | 66.19 | 53.87 | 62.06 | 56.70 |
| BERT$_{base}$-flow♡ | 58.40 | 67.10 | 60.85 | 75.16 | 71.22 | 68.66 | 64.47 | 66.55 |
| BERT$_{base}$-whitening♡ | 57.83 | 66.90 | 60.90 | 75.08 | 71.31 | 68.24 | 63.73 | 66.28 |
| IS-BERT$_{base}$♣ | 56.77 | 69.24 | 61.21 | 75.23 | 70.16 | 69.21 | 64.25 | 66.58 |
| CMLM-BERT$_{base}$♠ | 58.20 | 61.07 | 61.67 | 73.32 | 74.88 | 76.60 | 64.80 | 67.22 |
| CT-BERT$_{base}$♡ | 61.63 | 76.80 | 68.47 | 77.50 | 76.48 | 74.31 | 69.19 | 72.05 |
| SG-OPT-BERT$_{base}$♯ | 66.84 | 80.13 | 71.23 | 81.56 | 77.17 | 77.23 | 68.16 | 74.62 |
| SimCSE-BERT$_{base}$♡ | 68.40 | 82.41 | 74.38 | 80.91 | 78.56 | 76.85 | **72.23** | 76.25 |
| ∗ DiffCSE-BERT$_{base}$ | 68.43 | **82.73** | 73.83 | 81.56 | 78.67 | **78.76** | 72.09 | 76.58 |
| ∗ CMLM-CSE-BERT$_{base}$ | **70.52** | 82.20 | **74.89** | **82.58** | **79.06** | 78.18 | 70.20 | **76.80** |
| RoBERTa$_{base}$ (first-last avg.)♡ | 40.88 | 58.74 | 49.07 | 65.63 | 61.48 | 58.55 | 61.63 | 56.57 |
| RoBERTa$_{base}$-whitening♡ | 46.99 | 63.24 | 57.23 | 71.36 | 68.99 | 61.36 | 62.91 | 61.73 |
| DeCLUTR-RoBERTa$_{base}$♡ | 52.41 | 75.19 | 65.52 | 77.12 | 78.63 | 72.41 | 68.62 | 69.99 |
| SimCSE-RoBERTa$_{base}$♡ | 70.16 | 81.77 | 73.24 | 81.36 | 80.65 | 80.22 | 68.56 | 76.57 |
| ∗CMLM-CSE-RoBERTa$_{base}$ | **70.65** | **81.94** | **74.35** | **81.85** | 80.54 | **80.42** | 68.32 | **76.87** |

**4.3.2 Ablation Study**

(1) Removal of Contrastive Loss

Table 2: Experimental results of different loss functions on the STS-B validation set

| loss-function | w/o MLM loss | w/o contrastive loss | None |
|---|---|---|---|
| STS-B | 81.26 | 37.92 | 83.76 |

In our model, there are two important loss functions, the contrastive loss and the conditional MLM loss. The contrastive loss can pull similar sentences closer and push dissimilar sentences apart, while the MLM loss can predict the masked words in a sentence based on the sentence embedding, thus enabling the sentence embedding to capture important word-level features. In other words, the contrastive loss focuses on the global information of the sentence, while the conditional MLM loss focuses on the local information of the sentence. The corresponding ablation experiment is shown in Table 2, where we use the STS-B validation set for testing. After removing the conditional MLM loss, the model degenerates to SimCSE, and the Spearman correlation coefficient drops by 2.5 percentage points. After removing the contrastive loss, the overall model Spearman correlation coefficient drops by 45.84 percentage points. This result confirms our hypothesis that only conducting sentence-level contrastive learning ignores the word-level features in the sentence, and adding the conditional MLM loss compensates for the information of word-level features in the sentence embedding, thus improving the overall performance of the sentence embedding.

(2) Different data augmentation methods

Table 3: Experimental results on the STS-B validation set with different data augmentation

| Augmentation | Word Repetition | drop one word | None |
|---|---|---|---|
| STS-B | 75.84 | 79.16 | 83.76 |

We used two data augmentation methods on the contrastive loss side, namely adding duplicate words (first tokenizing the text and then repeating 32% of the words) and randomly deleting one word (first tokenizing the text and then randomly deleting one word). On the conditional MLM loss side, we still used random word masking of 15%. The specific results are shown in Table 3. On the STS-B validation set, the model without data augmentation had an accuracy of 83.76%, the accuracy with adding duplicate words was 75.84%, which was 7.92 percentage points lower than that without data augmentation, and the accuracy with randomly deleting one word was 79.16%, which was 4.6 percentage points lower than that without data augmentation.

(3) Masking rate

Table 4: Experimental results on the STS-B validation set with different word masking rates

| Rate | 15% | 20% | 25% | 30% | 40% | 45% |
|---|---|---|---|---|---|---|
| STS-B | 83.76 | 81.91 | 79.93 | 82.59 | **84.09** | 82.31 |

In the conditional MLM loss, we mask the original sentence with different ratios. The results are shown in Table 4. Different masking rates cause significant differences in performance. The 40% masking rate resulted in the highest validation set results, which were 4.16 percentage points higher than the lowest 25% masking rate.

(4) λ coefficient

Table 5: Experimental results on the STS-B validation set with different values of λ

| λ | 0 | 0.0001 | 0.0005 | 0.001 | 0.005 | 0.01 | 0.05 | 0.1 |
|---|---|---|---|---|---|---|---|---|
| STS-B | 81.26 | 82.45 | 81.82 | 83.55 | 83.76 | 82.48 | 81.52 | 81.96 |

We use the lambda coefficient to weight the conditional MLM loss and then add it to the contrastive loss. Because contrastive learning is a relatively simple task, we need a relatively small λ to balance the two losses. Table 5 shows the experimental results on the STS-B validation set with different λ. The best results were obtained when λ was 0.005. When λ is 0, the model degenerates into SimCSE.

(5) Auxiliary Network

Table 6: Performance results on the corresponding datasets with different numbers of layers for the lexical feature extractor and semantic feature fusion layer, using the STS-B test set for evaluation. (The far-right column represents the corresponding number of layers for the lexical feature extractor and semantic feature fusion layer, and the results are in terms of Spearman correlation coefficient.)

| Model Encoder Layers - Decoder Layers | STS12 | STS13 | STS14 | STS15 | STS16 | STS-B | SICK-R | Avg | STS-B (Dev) |
|---|---|---|---|---|---|---|---|---|---|
| 5-2 | 67.78 | 82.70 | 73.60 | 82.17 | 78.55 | 76.71 | 70.97 | 76.07 | 82.11 |
| 6-2 | 68.74 | 81.94 | 73.18 | 80.72 | 78.52 | 77.29 | 70.55 | 75.85 | 82.45 |
| 6-3 | 70.22 | 81.44 | 73.11 | 80.27 | 77.97 | 77.05 | **71.59** | 75.95 | 82.59 |
| 6-4 | **70.78** | 81.32 | 74.10 | 82.29 | 78.55 | 76.43 | 70.85 | 76.33 | 81.10 |
| 7-2 | 69.05 | 81.28 | 73.35 | 81.54 | 76.91 | 77.08 | 70.11 | 75.62 | 83.09 |
| 7-3 | 67.69 | 81.83 | 71.63 | 80.47 | 77.95 | 76.06 | 70.41 | 75.15 | 82.05 |

| | | | | | | | | |
|---|---|---|---|---|---|---|---|---|
| 8-2 | 68.41 | 80.02 | 73.42 | 81.35 | 77.42 | 76.00 | 68.39 | 75.00 | 82.58 |
| 8-3 | 70.52 | **82.20** | **74.89** | 82.58 | **79.06** | **78.18** | 70.20 | **76.80** | **83.76** |
| 8-4 | 67.72 | 79.56 | 72.07 | 79.38 | 77.77 | 76.52 | 71.15 | 74.88 | 81.43 |

The proportion of the auxiliary network (number of layers in the lexical feature extractor and semantic feature fusion block) is also crucial for the overall encoding performance. When the lexical feature extractor has more layers, it can provide better syntactic features, and the masked words can still be well reconstructed without relying on the sentence embeddings. When the number of layers in the lexical feature extractor is fewer, the information provided is less, or it may be biased towards lexical and surface features. The semantic feature fusion block needs more training to integrate this information and obtain syntactic and semantic features. Therefore, as shown in Table 6, when the encoder has fewer layers {<=6}, increasing the number of layers in the semantic feature fusion block can improve the average performance of the model on various tasks.

We conducted a series of ablation experiments under the condition of $\lambda = 0.005$ and the results are shown in Table 6. When the number of layers in the lexical feature extractor is 8 and the number of layers in the semantic feature fusion block is 3, the performance is the best, with a Spearman correlation coefficient of 83.76%. When the number of layers in the lexical feature extractor is 8 and the number of layers in the semantic feature fusion block is 2, the ability of the semantic feature fusion block is weak, and the model convergence is poor. The performance is 1.18% lower than when the number of layers in the semantic feature fusion block is 3. When the number of layers in the lexical feature extractor is 8 and the number of layers in the semantic feature fusion block is 4, too much semantic information is learned by the semantic feature fusion block, and the lexical feature extractor contains less semantic information, resulting in a performance 2.33% lower than when the number of layers in the semantic feature fusion block is 3.

## 5. conclusion

In this paper, we proposed CMLM-CSE, a contrastive learning framework based on conditional MLM, with an additional auxiliary network for MLM tasks. This auxiliary network fuses sentence embeddings to reconstruct the masked words, forcing the sentence embeddings to carry more information, thus called conditional masked language model. Our model achieved an average improvement of 0.55 percentage points over SimCSE in textual similarity tasks. We also conducted extensive ablation experiments to demonstrate the correctness and effectiveness of our approach.

## Acknowledgements

This research was supported by the Leading Innovative and Entrepreneurial Team of Hangzhou City (201920110028) and the Major Scientific and Technological Innovation Project on Artificial Intelligence of Hangzhou City (2022AIZD0093). We acknowledge the generous support of both funds, which has been instrumental to the completion of this work.